\documentclass[journal]{IEEEtran}
\usepackage{xcolor,soul,framed} 

\colorlet{shadecolor}{yellow}
\usepackage[pdftex]{graphicx}

\graphicspath{{../pdf/}{../jpeg/}}
\DeclareGraphicsExtensions{.pdf,.jpeg,.png}

\usepackage[cmex10]{amsmath}
\usepackage{array}
\usepackage{mdwmath}
\usepackage{mdwtab}
\usepackage{eqparbox}
\usepackage{amsmath}
\usepackage{amsfonts}
\usepackage{url}
\usepackage{subfigure}
\usepackage{caption}
\usepackage{multirow}
\usepackage{breqn}
\usepackage{lipsum}
\usepackage{booktabs}
\usepackage{verbatim}
\usepackage{mathtools}
\usepackage{hyperref}

\newcommand{\beq}{\begin{equation}}
\newcommand{\eeq}{\end{equation}}

\renewcommand{\vec}[1]{\mathbf{#1}}

\newcommand{\mr}[1]{\mathrm{#1}}

\newcommand{\yy}{\vec{y}}
\newcommand{\xx}{\vec{x}}
\newcommand{\zz}{\vec{z}}
\newcommand{\EE}{\mathbb{E}}

\newcommand{\Real}{\mathbb{R}}

\newcommand{\tsim}{\raise.17ex\hbox{$\scriptstyle\sim$}}

\bstctlcite{IEEE:BSTcontrol}

\begin{document}

\title{Few-shot 3D Multi-modal Medical Image Segmentation using Generative Adversarial Learning}

\author{Arnab Kumar Mondal, Jose Dolz and Christian Desrosiers
 \thanks{This work was being carried out in LIVIA, ETS Montreal.}
 \thanks{A. Mondal is an undergraduate student at IIT Kharagpur, India (e-mail: sanu.arnab@gmail.com).}
 \thanks{J. Dolz and C. Desrosiers are with the \'Ecole de technologie Superieure (ETS), Montreal, Canada.}%
 \thanks{}
 } 


\maketitle

\begin{abstract}
We address the problem of segmenting 3D multi-modal medical images in scenarios where very few labeled examples are available for training. Leveraging the recent success of adversarial learning for semi-supervised segmentation, we propose a novel method based on Generative Adversarial Networks (GANs) to train a segmentation model with both labeled and unlabeled images. The proposed method prevents over-fitting by learning to discriminate between true and fake patches obtained by a generator network. Our work extends current adversarial learning approaches, which focus on 2D single-modality images, to the more challenging context of 3D volumes of multiple modalities. The proposed method is evaluated on the problem of segmenting brain MRI from the iSEG-2017 and MRBrainS 2013 datasets. Significant performance improvement is reported, compared to state-of-art segmentation networks trained in a fully-supervised manner. In addition, our work presents a comprehensive analysis of different GAN architectures for semi-supervised segmentation, showing recent techniques like feature matching to yield a higher performance than conventional adversarial training approaches. Our code is publicly available at \url{https://github.com/arnab39/FewShot_GAN-Unet3D}.
\end{abstract}


%


\section{Introduction}

Semantic segmentation is commonly used in medical imaging to identify the precise location and shape of structures in the body, and is essential to the proper assessment of medical disorders and their treatment. Although extensively studied, this problem remains quite challenging due to the noise and low-contrast in medical scans, as well as the high variability of anatomical structures. Recently, deep convolutional neural networks (CNNs) have led to substantial improvements for numerous computer vision tasks like object detection \cite{ren2015faster}, image classification \cite{simonyan2014very,he2016deep} and semantic segmentation \cite{litjens2017survey,dolz20173d}, often achieving human-level performance. Yet, a major limitation of CNNs is their requirement for large amount of annotated data. This limitation is particularly important in medical image segmentation, where the annotation process is time-consuming, and prone to errors or intra-observer variability.

Semi-supervised learning approaches alleviate the need for large sets of labeled samples by exploiting available non-annotated data. In such approaches, only a limited number of samples with strong annotations are provided. A good generalization can however be achieved by considering unlabeled samples, or samples with weak annotations like image-level tags \cite{pinheiro2015weakly,papandreou2015weakly,kervadec2018constrained,pathak2015constrained}, bounding boxes \cite{dai2015boxsup,rajchl2017deepcut} or scribbles \cite{lin2016scribblesup,kervadec2018constrained}, during training. Recently, approaches based on adversarial training, and in particular Generative Adversarial Networks (GANs) \cite{NIPS2014_5423}, have shown great potential for improving semantic segmentation in a semi-supervised setting \cite{souly2017semi,Hung_semiseg_2018}. Nevertheless, their application to the segmentation of 3D medical images with multiple modalities remains to this day limited \cite{rezaei2017conditional,li2017brain}.

Our work addresses the problem of segmenting 3D multi-modal medical images from a few-shot learning perspective. We leverage the recent success of GANs to train a deep model with a highly-limited training set of labeled images, without sacrificing the performance of full supervision. The main contributions of this paper can be summarized as:
\begin{enumerate}\setlength\itemsep{0.4em}
\item A first approach to apply GANs for the semi-supervised segmentation of 3D multi-modal images. We demonstrate this approach to significantly outperform state-of-art segmentation networks like 3D-UNet when very few training samples are available, and to achieve an accuracy close to that of full supervision. 
\item A comprehensive analysis of different GAN architectures for semi-supervised segmentation, where we show more recent techniques like feature matching to have a higher performance than conventional adversarial training approaches. 
\end{enumerate}

The rest of this paper is organized as follows. In Section \ref{sec:related_works}, we give a brief overview of relevant work on semantic segmentation with a focus on semi-supervised learning. Section \ref{sec:methods} then presents our 3D multi-modal segmentation approach based on adversarial learning, which is evaluated on the challenging task of brain segmentation in Section \ref{sec:experiments}. Finally, we conclude with a summary of our main contributions and results.

\begin{figure*}[ht!]
 \centering
 \shortstack{
 \includegraphics[width=0.75\linewidth]{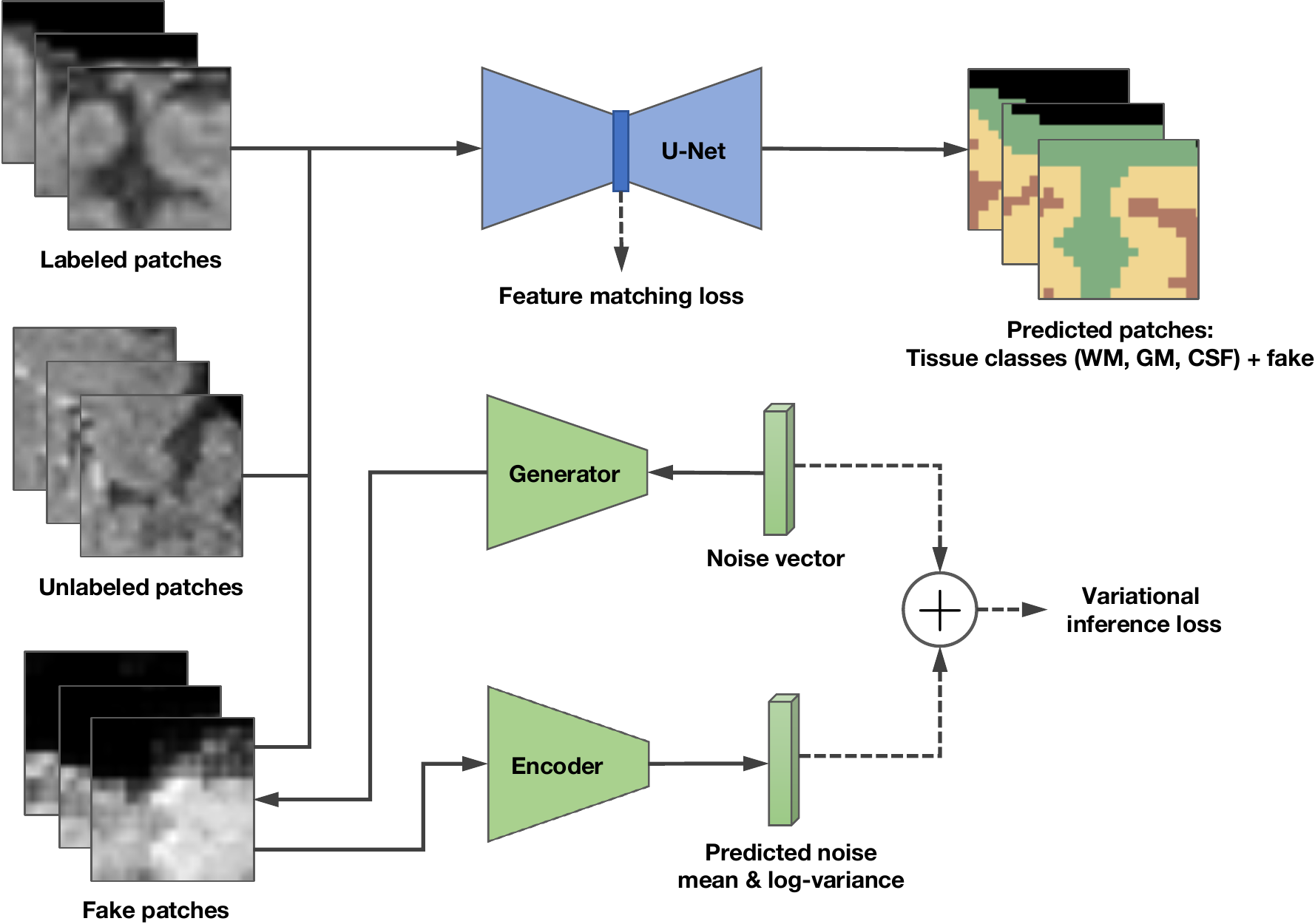}
 \\ \vphantom{1mm}}
 \caption{Schematic explaining the working of our model. The model contains three networks which are trained simultaneously.}
 \label{fig:architecture}
\end{figure*}

\section{Related work}\label{sec:related_works}

Our method draws on recent successes of deep learning methods for semantic segmentation, in particular semi-supervised and few-shot learning approaches based on adversarial training.

\subsection{Semi-supervised learning}

Several semi-supervised deep learning methods have been proposed for image segmentation \cite{bai2017semi,baur2017semi,min2018robust,zhou2018semi}. A common strategy, based on the principle of self-training, involves updating network parameters and segmentation alternatively until convergence \cite{bai2017semi}. However, if initial class priors given by the network are inaccurate, segmentation errors can occur and be propagated back to the network which then re-amplifies these errors. Various techniques can be used to alleviate this problem, including model-based \cite{gupta2016cross} or data-based \cite{radosavovic2017data,zhou2018semi} distillation, which aggregate the prediction of multiple teacher models or a single teacher trained with multiple transformed versions of the data to learn a student model, and employing attention modules \cite{min2018robust}. Yet, these approaches are relatively complex, as they require to train multiple networks, and are thus not suitable when very few training samples are available. Another popular approach consists in embedding the network's output or internal representation in a manifold space, such that images having similar characteristics are near to each other \cite{baur2017semi}. An important limitation of this approach is its requirement for an explicit matching function, which may be hard to define in practice. 

\subsection{Adversarial learning}

Adversarial learning has also shown great promise for training deep segmentation models with few strongly-annotated images \cite{souly2017semi,Hung_semiseg_2018,zhang2017deep,luc2016semantic}. An interesting approach to include unlabeled images during training is to add an adversarial network in the model, which must determine whether the output of the segmentation network corresponds to a labeled or unlabeled image \cite{zhang2017deep,luc2016semantic}. This encourages the segmentation network to have a similar distribution of outputs for images with and without annotations, thereby helping generalization. A potential issue with this approach is that the adversarial network can have a reverse effect, where the output for annotated images becomes growingly similar to the incorrect segmentations obtained for unlabeled images. A related strategy uses the discriminator to predict a confidence map for the segmentation, enforcing this output to be maximum for annotated images \cite{Hung_semiseg_2018}. For unlabeled images, areas of high confidence are used to update the segmentation network in a self-teaching manner. The main limitation of this approach is that a confidence threshold must be provided, the value of which can affect the performance. Up to date, only a single work has applied Generative Adversarial Networks (GANs) for semi-supervised segmentation \cite{souly2017semi}. However, it focused on 2D natural images, whereas the current work targets 3D multi-modal medical volumes. Generating and segmenting 3D volumes brings additional challenges, such as computational complexity and over-fitting.


\subsection{Few-shot learning}

Few-shot learning methods seek good generalization on problems with a very limited labeled dataset, typically containing just a few training samples of the target classes \cite{fei2006one,vinyals2016matching}. Even tough interest in such methods is increasing, most works focus on classification \cite{ren2018meta}, where there is no need to generate a structured output. Shaban et al. \cite{shaban2017one} pioneered one-shot learning for semantic segmentation, which required only a single image and its corresponding pixel-level annotation per class, for learning. More recently, Rakelly et al. proposed to train guided \cite{rakelly2018few} and conditional \cite{rakelly2018conditional} networks with few support samples and sparse annotations, achieving similar performance to \cite{shaban2017one}. A common feature of these methods is that a network pre-trained on a similar task is employed as initial model, which is then refined by the one/few shot support queries. However, adopting this strategy to segment medical images might be challenging as the gap between source and target tasks is often large. As shown in our experiments, the approach proposed in this work can achieve competitive performance using as few as two training samples, without the need for a pre-trained model.

\section{Methods}\label{sec:methods}

The proposed architecture for the semi-supervised segmentation of 3D medical images is illustrated in Figure \ref{fig:architecture}. In a standard segmentation model such as U-Net \cite{ronneberger2015u}, fully-annotated images are typically employed to train the network using a pixel-wise loss function like cross-entropy. As mentioned above, this is not possible in our case since the number of annotated images for training is highly limited. As in other semi-supervised segmentation approaches, we alleviate this problem by also incorporating unlabeled images in the training process. However, unlike these methods, we also make use of synthetic (i.e., \emph{fake}) images generated by a GAN.

To include labeled, unlabeled and fake images during training, we extend the classification approach of Dai et al. \cite{NIPS2017_7229} to segmentation. In this GAN-based approach, a generator network is used to produce realistic fake examples and a discriminator to distinguish these fake examples from true data. Instead of predicting $K$ classes, as in standard methods, the model predicts ($K$+1) classes, where the additional class corresponds to fake examples. However, as we show in following subsections, this formulation can be recast back into a $K$-class problem using a simple re-parametrization trick. This overall strategy helps the model give plausible predictions for unlabeled true data by restricting its output for fake examples.

For adapting this model to the segmentation of 3D multi-modal images, several changes must be made. During training, 3D images must be processed in smaller sub-regions (i.e., \emph{patches}) to deal with the much greater memory and computational requirements compared to 2D images. While training patches have a fixed size, test images may have arbitrary size. To address this issue, we make the segmentation network to be fully-convolutional \cite{long2015fully}. Another challenge comes from the generation of fake patches. Standard techniques for training GANs may lead to instability and poor results, especially in the case of semi-supervised learning \cite{NIPS2016_6125}. This problem is even more significant in the case of 3D multi-modal patches, whose distribution is harder to estimate with a parametric model. In addition, although generated patches must be realistic-looking, they should be sufficiently different from true unlabeled patches, otherwise the wrong information will be learned \cite{NIPS2017_7229}. 

The following subsections provide a more detailed description of the proposed method. We start by giving a general formulation of generative adversarial networks (GANs). Then, we show how GANs can be used to include unlabeled and fake images in a semi-supervised segmentation setting. Finally, we explain how the standard GAN model is modified to fit our problem setting.       

\subsection{Generative Adversarial Networks}

Standard GANs \cite{NIPS2014_5423} model a min-max game between two parametrized neural nets: a generator $G_{\theta_G}$ and a discriminator $D_{\theta_D}$. The idea is to train the generator to generate images from a learned data distribution $p_{\mr{data}}(\xx)$, while simultaneously training the discriminator to differentiate between these generated images and true examples. Specifically, $G_{\theta_G}$ is trained to map a random (i.e., \emph{noise}) vector $\zz \in \Real^d$ to a synthetic image vector $\tilde{\xx} = G(\zz)$. A common choice for sampling the noise is to use a uniform distribution, e.g., $\zz \sim \mathcal{U}[-1,1]^d$. 
On the other hand, $D_{\theta_D}$ is trained to distinguish between real examples $\xx \sim p_{\mr{data}(\xx)}$ and synthetic examples $\tilde{\xx} \sim p_{G(\zz)}$. Here, $D_{\theta_D}(\xx)$ represents the probability that a sample $\xx$ belongs to original data distribution. Networks $D$ and $G$ play a two player min-max game with value function $V(D_{\theta_D}, G_{\theta_G})$:
\begin{equation}\label{eq_gan_main_goodfellow}
\begin{aligned}
\min_{G_{\theta_G}} \ \max_{D_{\theta_D}} & \ \ 
	 \EE_{\xx\sim p_{\mr{data}}(\xx)}\,\big[\log D_{\theta_D}(\xx)\big]\\
& \quad \quad + \, \EE_{\zz\sim \mr{noise}}\,\big[1 - D_{\theta_d}(G_{\theta_G}(\zz))\big]
\end{aligned}
\end{equation}


\subsection{Reduced supervision with GANs}

Consider a standard CNN-based model for segmenting a 3D image $\xx_{H\times W\times D}$ into regions defined by $\yy_{H\times W\times D}$. This model takes $\xx$ as input and outputs a $K$-dimensional vector of logits $\left [ l_{i,1}, \ldots , l_{i,K} \right ]$, where $K$ is the number of classes labels and $i$ is the index of image voxels. This output can be turned into class probabilities by applying the softmax function: 
\begin{equation}\label{eq:softmax}
	p_{\mr{model}}(y_{i}=j \, | \, \xx) \, = \, 			
    	\frac{\exp(l_{i,j})}{\sum_{k=1}^{K} \exp(l_{i,k})}.
\end{equation}
In a fully-supervised setting, the model is typically trained by minimizing a segmentation loss function, for instance, the cross-entropy between the true labels and the model's predicted probabilities.

As shown in Fig. \ref{fig:architecture}, the proposed model extends the standard full-supervision approach by incorporating unlabeled data and samples from the generator $G$ during training. Toward this goal, we label generated samples with a new class $y_i=K\!+\!1$ and thus increase the dimension of the segmentation model's output to $H\!\times\!W\!\times\!D\!\times\!(K\!+\!1)$. In this new formulation, $p_{\mr{model}}(y_i=K+1\, | \,\xx)$ is the probability that voxel $i$ of input $\xx$ is fake. Moreover, to learn the basic structure of images from unlabeled data, we constraint the output to correspond to one of the $K$ classes of real data, which can be done by maximizing
\begin{equation}\label{eq:map_unlabeled}
\EE_{\xx\sim p_{\mr{data}}(\xx)} \sum_{i=1}^{H\times W\times D} \!\!\!\!\log \, p_{\mr{model}}(y_i \in \left \{ 1,\ldots,K \right \}\, | \,\xx).
\end{equation}
With this, we can now define the loss functions used for training the discriminator and generator networks.

\subsubsection{Discriminator loss}

Suppose have a similar number of labeled, unlabeled and fake images, so that each type of images has equal importance in training. Our discriminator loss function can be defined as the sum of three terms: 
\begin{equation}\label{eq_total_loss}
L_{\mr{discriminator}} \, = \, L_{\mr{labeled}} \, + \, L_{\mr{unlabeled}} \, + \, L_{\mr{fake}}.
\end{equation}
The loss for labeled images is the same as in standard segmentation networks. In this work, we consider the mean cross-entropy:
\begin{equation}\label{eq:loss_labeled}
\begin{aligned}
& L_{\mr{labeled}} \, = \\[0mm] 
& \ -\,\EE_{\xx,\yy\sim p_{\mr{data}}(\xx,\yy)} \!\!\sum_{i=1}^{H\times W\times D}
	\!\!\!\!\!\log \, p_{\mr{model}}(y_i\, | \,\xx, \, y_i\!<\!K\!+\!1).
\end{aligned}
\end{equation}
In the case of unlabeled images, we maximize the term in Eq.~(\ref{eq:map_unlabeled}), which is the same as minimizing
\begin{equation}\label{eq:loss_unlabeled}
\begin{aligned}
& L_{\mr{unlabeled}} \, = \\[0mm] 
	& - \, \EE_{\xx\sim p_{\mr{data}}(\xx)} \!\! \sum_{i=1}^{H\times W\times D} \!\!\!\!\!\log \, \big[1 - p_{\mr{model}}(y_i = K\!+\!1 \, | \,\xx)\big]
\end{aligned}
\end{equation}
Finally, for generated images, we impose each pixel of an input patch to be predicted as fake, and define the loss as 
\begin{equation}\label{eq:loss_fake}
\begin{aligned}
& L_{\mr{fake}} \, = \, - \, \EE_{z\sim \mr{noise}} \!\! \sum_{i=1}^{H\times W\times D} \!\!\!\!\!\log \, p_{\mr{model}}\big(y_i = K\!+\!1 \, | \, G_{\theta_G}(z)\big)
\end{aligned}
\end{equation}

In \cite{NIPS2016_6125}, it was shown that the optimal strategy for minimizing Eq.~(\ref{eq_total_loss}) is to have $\exp[l_{i,j}] = c_i(\xx) \cdot p(y_i=j, \,\xx), \quad \forall j<K+1$ and $\exp[l_{i,K+1}] = c_i(\xx) \cdot p_G(\xx)$, where $c_i(\xx)$ is an undetermined scaling function for the $i$-th pixel. 
It was also found that the having $K+1$ outputs is an over-parameterized formulation, since subtracting a general function $f(\xx)$ from each logit does not change the output of the softmax. By using the logit of the fake class $l_{i,K+1}$ as substracted function, we  obtain $\big[(l_{i,1} - l_{i,K+1}), \ldots, (l_{i,K} - l_{i,K+1}), 0\big]$, and thus have only $K$ effective (i.e., non-zero) outputs. Employing these ``normalized'' logits in the softmax of Eq.~(\ref{eq:softmax}) then leads to 
the following modified loss functions: 
\begin{align}\label{eq_losses}
L_{\mr{labeled}} & \, = \, - \, \EE_{\xx,\yy\sim p_{\mr{data}}(\xx,\yy)} \!\!\sum_{i=1}^{H\times W\times D}\!\!\!\log \, p_{\mr{model}}(y_i \, | \,\xx) \\[6pt]
L_{\mr{unlabeled}} & \, = \, - \, \EE_{\xx\sim p_{\mr{data}}(\xx)}  \!\!\sum_{i=1}^{H\times W\times D}
	\!\!\!\log \left[\frac{Z_i(\xx)}{Z_i(\xx)+1}\right] \\[6pt]
L_{\mr{fake}} & \, = \, - \, \EE_{\zz\sim \mr{noise}} \!\!\sum_{i=1}^{H\times W\times D} 
	\!\!\!\log \left[\frac{1}{Z_i(G_{\theta_G}(\zz))+1}\right]
\end{align}
where $Z_i(x) = \sum_{k=1}^{K}\exp[l_{i,k}(\xx)]$.

In summary, the idea is to plug a standard state-of-the-art segmentation model in the discriminator of the proposed network, where the labeled component of the loss $L_{\mr{labeled}}$ remains unchanged (i.e, cross-entropy), and introduce two extra terms, the unlabeled term $L_{\mr{unlabeled}}$ and the fake term $L_{\mr{fake}}$, which are analogous to the two components of a discriminator loss in standard GANs. 

\subsubsection{Generator loss}

The most common strategy for training the generator consists in maximizing the $L_{\mr{fake}}$ loss of Eq.~(\ref{eq:loss_fake}). However, as demonstrated in \cite{NIPS2016_6125}, this can lead to instability and poor performance in the case of semi-supervised learning. Following these results, we instead adopt the Feature Matching (FM) loss for the generator, which is more suited to our problem. In FM, the goal of the generator is to match the expected value of features $f(\xx)$ in an intermediate layer of the discriminator:
\begin{equation}\label{eq_feature_matching}
L_{\mr{generator}} \, = \, \big\|E_{\xx\sim p_{\mr{data}}(\xx)}\, f(\xx) 
	\, - \, E_{z \sim \mr{noise}} \, f(G_{\theta_G}(\zz))\big\|_{2}^{2}
\end{equation}
In this work, $f(\xx)$ contains the activations of the second last layer of the encoding path in our model. In preliminary experiments, we found this choice to give slightly higher performance than using the encoder's last layer.

\subsection{Complement (Bad) generator}\label{sec:bad-gan}


In semi-supervised learning, having a good generator can actually deteriorate performances since in this case unlabeled and fake images cannot be separated. It is therefore desirable to have a generator that can generate samples outside the true data manifold, which is called a \emph{complement} (or \emph{bad}) generator \cite{NIPS2017_7229}.

The FM generator loss, described in the previous section, works better than standard training approaches in a semi-supervised setting because it performs distribution matching in a weak manner. However, it may still face two significant problems.
First, since an FM-based generator can assign a significant amount of probability mass inside the support, an optimal discriminator will incorrectly predict samples in that region as fake. Secondly, as FM only matches first-order statistics, the generator might end up with a trivial solution, for example, it can collapse to mean of unlabeled features. The collapsed generator will then fail to cover some areas between manifolds. Since the discriminator is only well-defined on the union of the data supports of $p$ and $p_{G}$, the prediction result in such gaps is under-determined.

The first problem is less likely in our case, since multi-modal 3D patches are complex structures to generate and, thus, it is more probable for the FM generator to sample images outside the true data manifold. 
To deal with the second problem, we can increase the entropy of the generated distribution by minimizing a modified loss for the generator: 
\begin{equation}
\begin{split}
L_{generator} \, = \, & -\mathit{H}(p_G) \, + \, \\ 
	& \big\| E_{\xx\sim p_{\mr{data}}(\xx)} \, f(\xx)\, - \, E_{\xx\sim p_G} \, f(\xx)\big\|_{2}^{2}
\end{split}
\label{eq_feature_matching}
\end{equation}
As mentioned in \cite{NIPS2017_7229}, this complex loss function can be optimized using a variational upper bound on the negative entropy \cite{dai2017calibrating}:
\begin{equation}
	-\mathit{H}\big(p_G(\xx)\big) \leq -~\EE_{\xx,\zz\sim p_G} \log q(\zz\, | \,\xx).
\end{equation}
In this formulation, $q$ is defined as a diagonal Gaussian with bounded variance, i.e. $q(\zz \, | \, \xx)= \mathcal{N}(\mu(\xx),\,\sigma^{2}(\xx))\,$ , with $0\leq \sigma(\xx) \leq \theta$, where $\mu(\cdot)$ and $\sigma(\cdot)$ are neural networks. 

The overall architecture of the complement generator is illustrated in Fig. \ref{fig:architecture}. As presented above, the FM loss uses features from the second layer of the discriminator (i.e., the U-Net segmentation network). Moreover, the fake image generator is paired with an encoder which learns a reverse mapping from generated images to corresponding noise vectors. All  components the architecture are trained simultaneously in an end-to-end manner.




\section{Experiments}\label{sec:experiments}

\subsection{Materials}

The proposed model is evaluated on the challenging tasks of segmenting infant and adult brain tissue from multi-modal 3D magnetic resonance images (MRI). The goal of our experiments is two-fold. First, we assess our GAN-based model in a few-shot learning scenario, where only a few training subjects are provided. Our objective is to provide performance similar to that of full-supervision, while using only 1 or 2 training subjects. Second, since the application of GANs to semi-supervised learning, and particularly to segmentation, is a new topic, we conduct experiments to measure to impact of various GAN techniques (e.g., feature matching, complementary generator, etc.) on segmentation accuracy. Before presenting results, we give details on the dataset, evaluation metrics and implementation used in the experiments.



\subsubsection{Dataset}

We first used data from the iSEG-2017\footnote{See \url{http://iseg2017.web.unc.edu/reference/}} Challenge on infant brain MRI segmentation. The goal of this challenge is to compare (semi-) automatic algorithms for the segmentation of infant (\tsim6 months) T1- and T2-weighted brain MRI scans into three tissue classes: white matter (WM), gray matter (GM) and cerebrospinal fluid (CSF). 
This dataset was chosen to substantiate our proposed method: it contains the 3D multi-modal brain MRI data of only 10 labeled subjects, each one requiring about a week to annotate manually. Additionally, 13 unlabeled testing subjects are also provided. To further validate results, we also tested our method on segmenting adult brain tissues (i.e., WM, GM, CSF) from the MRBrains-2013\footnote{See \url{http://mrbrains13.isi.uu.nl}} Challenge dataset, which contains the T1, T1-IR and FLAIR scans of 20 adult subjects. Ground truth labels are provided for only 5 training subjects, which form the training set. The test set contains the unlabeled scans of the 15 remaining subjects.


\subsubsection{Evaluation}

Segmentation accuracy is assessed using two well-known metrics, respectively measuring spatial overlap and surface distance \cite{yeghiazaryan2015overview}:
\begin{itemize}\setlength\itemsep{0.4em}
\item \textit{Dice similarity coefficient} (DSC): This widely-used metric compares segmented volumes based on their overlap. Given a reference segmentation $S_{\mr{ref}}$, the DSC of a predicted segmentation $S_{\mr{pred}}$ is defined as \begin{equation} 
\mathrm{DSC} \ = \ \dfrac{2\left|S_{\mr{pred}} \cap S_{\mr{ref}} \right|}{|S_{\mr{pred}}| \,+ \, |S_{\mr{ref}}|}.
\end{equation}
DSC values range between 0 and 1, with 1 corresponding to a perfect overlap.
\item \textit{Average Symmetric Surface Distance} (ASD): This metric computes an average of distances from points on a surface to the nearest point on another surface. Let $B_{\mr{ref}}$ and $B_{\mr{pred}}$ be the reference and predicted segmentation boundaries, it can be defined as
\begin{equation}
ASD \, = \, \frac{1}{N}
	\left(\!\!\!\sum_{~~x \in B_{\mr{pred}}} \!\!\!\!\!\!d(x,B_{\mr{ref}})
		 \, + \!\!\!\!\sum_{~~x\in B_{\mr{ref}}} \!\!\!\!d(x, B_{\mr{pred}})\right),
\end{equation}
where $N=|B_{\mr{pred}}| + |B_{\mr{ref}}|$.
\end{itemize}


\subsection{Implementation Details}

\subsubsection{Architecture} \label{sec:u-net}

The state-of-art 3D U-Net \cite{cciccek20163d} model was chosen as segmentation network in our architecture. In order to use this model in the proposed GAN framework, the following changes were made:
\begin{itemize}\setlength\itemsep{0.4em}
\item Batch-normalization \cite{ioffe2015batch} was replaced by weight-normalization \cite{salimans2016weight}, since the former had detrimental effect on GAN training for semi-supervised learning.
\item As suggested in \cite{NIPS2016_6125}, ReLUs were changed to leaky ReLUs, allowing a small gradient for non-active units (i.e., units whose output is below zero).
\item Max pooling was replaced by average pooling, as it leads to sparse gradient which was shown to hamper GAN training. 
\end{itemize} 
These modifications to 3D U-Net have helped make the training more stable and improve the performance. Other elements of the discriminator's architecture are the same as in the original U-Net. 

For generating 3D patches, we chose the volume generator proposed by Wu et al. \cite{wu2016learning}, which was shown to provide good results for various types of 3D objects. This model leverages the power of both general-adversarial modeling and volumetric convolutional networks to generate realistic 3D shapes. For implementing the encoder, we used a standard three-layer 3D CNN architecture, whose output vector is twice the size of the generator's input noise vector. This network estimates the mean and standard deviation of the noise vector from which the given image is generated. It was found during preliminary experiments that using batch normalization in the generator and encoder gives best results. Therefore, this normalization setting was used for our GAN-based model.

\subsubsection{Training}

  
To train the proposed GAN based model, the 10 labeled subjects data (i.e., examples) of the iSEG-2017 dataset were split into training (1 or 2 examples), validation (1 example) and testing (7 fixed examples). The 13 unlabeled examples of the testing dataset were instead used to train the GAN. 

Similarly, for the MR Brains 2013 dataset, the 5 labeled examples were split into 1 training, 1 validation and 3 testing examples, respectively. As before, the 15 unlabeled subject data were used as unlabeled data for training the GAN. 

As preprocessing, N4 bias field correction was applied to images, followed by intensity normalization. To train the model, 32$\times$32$\times$32 patches were extracted from 3D scans with a step size of 8 voxels in each dimension. This serves two purposes: reduce computational requirements compared to employing whole 3D images, and increase the number and diversity of training examples. No other data augmentation was used, as our goal is to compare the performance of the two models in a few-shot learning scenario, not to achieve state-of-the-art performance on the tested datasets. The Adam optimizer was employed for mini-batch stochastic gradient descent (SGD), with a batch size of 30. For all networks (i.e., U-Net based discriminator, generator and encoder), we used a learning rate of 0.0001 and a momentum of 0.5. 



%
\begin{table*}[ht!]
\centering
\caption{DSC and ASD (mm) computed on 7 test examples of the iSEG-2017 dataset, using either 1 or 2 labeled examples for training and 1 example for validation. Reported results are the average over three different splits of the training and validation examples. For our GAN-based model, the 13 unlabeled examples of the dataset are also employed during training. Best results are highlighted in bold}
\renewcommand{\arraystretch}{1.2}
\begin{tabular}{lcllcllclccllcllcl}
\toprule
  & \multicolumn{8}{c}{\textbf{\emph{1 training image}}}    & & \multicolumn{8}{c}{\textbf{\emph{2 training images}}}    \\ 
  \cmidrule(l{5pt}r{5pt}){2-9} \cmidrule(l{5pt}r{5pt}){11-18} 
  \multirow{2}{*}{\textbf{Method used}} & \multicolumn{2}{c}{\textbf{WM}} & \textbf{} & \multicolumn{2}{c}{\textbf{GM}} & \textbf{} & \multicolumn{2}{c}{\textbf{CSF}} & \textbf{} & \multicolumn{2}{c}{\textbf{WM}} & \textbf{} & \multicolumn{2}{c}{\textbf{GM}} & \textbf{} & \multicolumn{2}{c}{\textbf{CSF}} \\ 
\cmidrule(l{5pt}r{5pt}){2-3}
\cmidrule(l{5pt}r{5pt}){5-6} 
\cmidrule(l{5pt}r{5pt}){8-9} 
\cmidrule(l{5pt}r{5pt}){11-12} 
\cmidrule(l{5pt}r{5pt}){14-15} 
\cmidrule(l{5pt}r{5pt}){17-18}
& \multicolumn{1}{l}{DSC} & ASD & & \multicolumn{1}{l}{DSC} & ASD & & \multicolumn{1}{l}{DSC} & ASD  & \multicolumn{1}{l}{} & \multicolumn{1}{l}{DSC} & ASD & & \multicolumn{1}{l}{DSC} & ASD & & \multicolumn{1}{l}{DSC} & ASD \\ 
\midrule\midrule
Basic U-Net & 0.61 & 1.89 & & 0.49 & 2.25 & & 0.80 & 0.60 & & 0.68 & 1.43 & & 0.62 & 1.79 & & 0.82 & 0.59 \\
\midrule
Ours (normal GAN) & 0.66 & 1.75 & & 0.62 & 1.91 & & 0.81 & 0.62 & & 0.71 & 0.96 & & 0.72 & 0.89 & & 0.82 & 0.51 \\
Ours (FM GAN) & \textbf{0.74} & \textbf{0.82} & \textbf{} & \textbf{0.72} & \textbf{0.85} & \textbf{} & \textbf{0.89} & \textbf{0.27} & \textbf{} & \textbf{0.80} & \textbf{0.54} & \textbf{} & \textbf{0.80} & \textbf{0.58} & \textbf{} & \textbf{0.88} & \textbf{0.25} \\
Ours (bad-GAN)
& 0.69 & 1.20 & & 0.68 & 1.33 & & 0.86 & 0.39 & & 0.74 & 0.69 & & 0.76 & 0.66 & & 0.84 & 0.41 \\ 
\bottomrule
\end{tabular}
\label{tab1}
\end{table*}

\begin{figure*}[ht!]
 \centering
 \subfigure[White Matter]{\includegraphics[width=55mm]{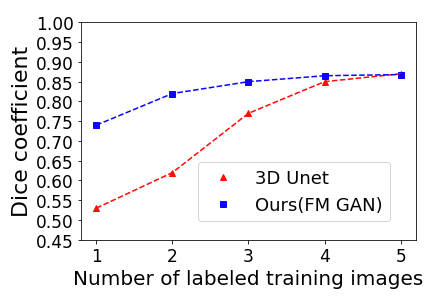}\label{fig61}}
 \subfigure[Gray Matter]{\includegraphics[width=55mm]{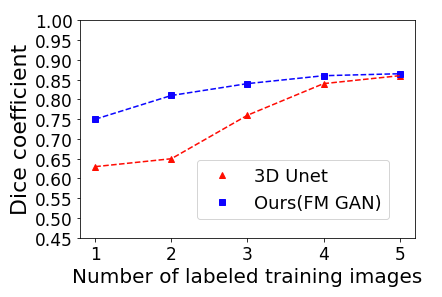}\label{fig62}}
 \subfigure[Cerebrospinal Fluid]{\includegraphics[width=55mm]{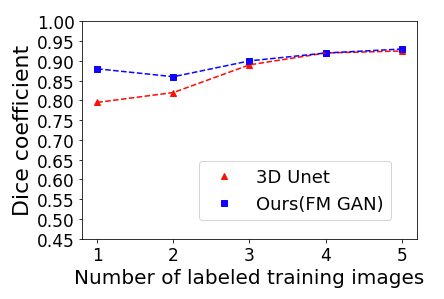}\label{fig63}}
 \caption{DSC of U-Net and our FM GAN model when training with an increasing number of labeled examples from the iSEG-2017 dataset. Performance is measured on 4 test examples, and a single validation example is used during training. For our GAN-based model, the 13 unlabeled examples of the dataset are also employed during training.}
 \label{fig6}
\end{figure*}

\subsection{Few-shot learning with FM GAN}

To validate the proposed model in a few-shot learning scenario, we trained it end-to-end with only 1 or 2 training examples. The objective is to show that, when training with few labeled examples, our model outperforms U-Net and gives performance close to full-supervision without data augmentation. 
While training, the model is validated with a single labeled example, thus making the total number of labeled examples no greater than 3. To reduce bias while estimating performance, we repeated this process with 3 different combinations of training and validation examples, while keeping the 7 test examples fixed, and report the average result.

Table \ref{tab1} gives the mean DSC and ASD obtained by the 3D U-Net modified as described in Section \ref{sec:u-net} (Basic U-Net), and our proposed model with standard adversarial loss (Normal GAN), feature matching (FM GAN), or the complementary GAN model of Section \ref{sec:bad-gan} (bad-GAN). Results are reported for 1 and 2 labeled training examples. We see that the proposed GAN-based method significantly outperforms basic U-Net when a single labeled example is available, with DSC improvements of 5-8\% for WM, 13-23\% for GM, and 1-9\% for CSF. Important improvements are also observed for 2 labeled examples, with a DSC increase of 3-12\%, 10-18\% and 1-6\% for WM, GM and CSF, respectively. Similarly, we see a significant reduction in ASD for both cases. 

Comparing the different GAN models, we find that feature matching (without entropy term) yields the best performance, for all tissue classes and test cases. Compared to bad-GAN, it provides DSC improvements of 5\% for WM, 4\% for GM and 3\% for CSF, in the case of 1 labeled example, and improvements of 6\% for WM, 4\% for GM and 4\% for CSF, when 2 labeled examples are employed. In the next section, we analyze in greater detail the behavior of these two GAN models to better understand these results.

Next, we evaluate the impact of supervision on the performance of 3D U-Net and FM GAN by increasing the number of labeled images in training from 1 to 5. Results of these experiments are plotted in Figure \ref{fig6}. In this experiment, we used a single validation example and a fixed set of 4 test examples. 
It can be seen that, compared to U-Net, FM GAN gives a higher or equal DSC in all cases, and that the accuracy of models is comparable for 5 labeled examples. Although 5 examples seems like a relatively small number, one should remember that networks are trained using patches sampled over these images, and thus these networks see thousands of training patches.


\begin{figure*}[ht!]
 \centering
 \mbox{
\includegraphics[width=0.48\linewidth]{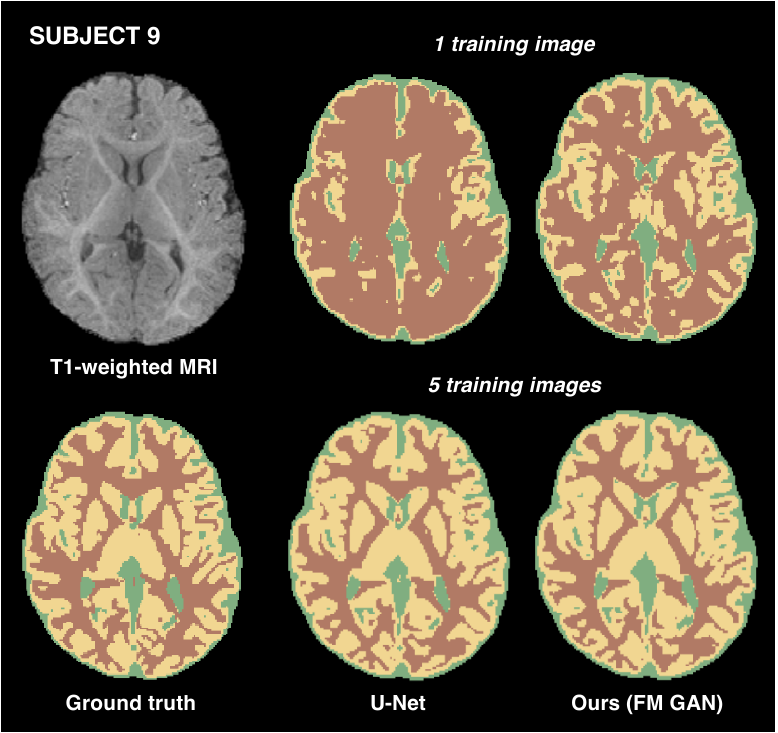} 
 
 \hspace{0.5mm}
 
 \includegraphics[width=0.48\linewidth]{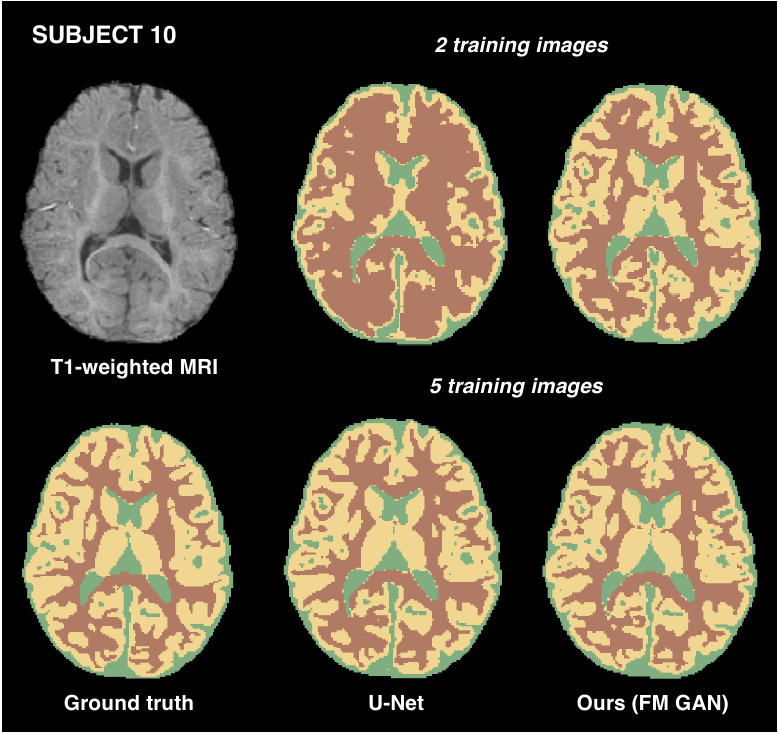}
 }
 \caption{Visual comparison of the segmentation by each model, for two test subjects of the iSEG-2017 dataset, when training with different numbers of labeled examples.}
 \label{fig:Unet_vs_FMGAN}
\end{figure*}

To visually appreciate the performance of the proposed model, Figure \ref{fig:Unet_vs_FMGAN} shows the segmentation output of Basic U-Net and FM GAN for two different subjects, when training with 1, 2 or 5 labeled examples. If 1 or 2 labeled examples are used, standard U-Net gives poor results, showing the inability of this model to work in a few shot learning scenario. In contrast, FM GAN can better learn the structure of brain tissues by using unlabeled images. Moreover, following the results of Fig. \ref{fig6}, we see that the segmentation of FM GAN is visually similar to U-Net when 5 labeled images are employed in training.  


\subsection{Detailed analysis of GAN models}

\begin{figure}[ht!]
 \centering
 \subfigure[U-Net]{\includegraphics[width=0.75\linewidth]{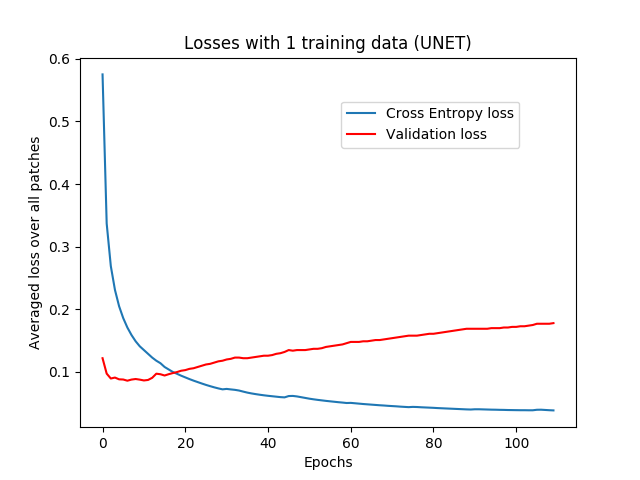}\label{fig71}}
 \subfigure[Ours (FM GAN)]{\includegraphics[width=0.75\linewidth]{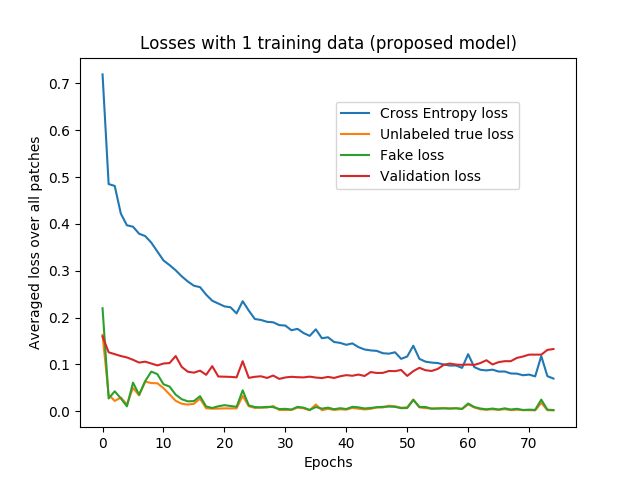}\label{fig72}}
 \caption{Loss of U-Net and our FM Gan model at different  training epochs, measured on a random subset of validation patches.}
 \label{fig9}
\end{figure}

Results of the previous experiment showed the proposed model to outperform standard U-Net when very few labeled images are provided in training. In this section, we try to explain how the unlabeled and fake components of the loss function enable such improvements. Moreover, we analyze the tested GAN models to determine which elements contribute to having accurate segmentations.  

Figure \ref{fig9} plots the training losses of U-Net and our FM GAN model, at different training epochs, when using a single labeled example. For U-Net, we show the cross-entropy loss of Eq.~(\ref{eq:loss_labeled}) and validation error (i.e., mean percentage of incorrectly predicted voxels in randomly selected patches of validation images). In the case of FM GAN, we also report the unlabeled image of Eq.~(\ref{eq:loss_unlabeled}) and fake image loss of Eq.~(\ref{eq:loss_fake}). These plots clearly show how U-Net, being a high-capacity model, quickly overfits the data. In contrast, our FM GAN model also learns from unlabeled and generated data and, hence, generalizes better the validation data.

\begin{table}[ht!]
\centering
\caption{Mean unlabeled and fake loss computed over patches extracted from test images, when training with 2 labeled examples. Best results highlighted in bold.}
\renewcommand{\arraystretch}{1.3}
\begin{tabular}{lcc}
\toprule
\multicolumn{1}{l}{\textbf{Method}} &  \textbf{Unlabeled loss} & \textbf{Fake loss} \\ 
\midrule\midrule
Basic U-Net & 0.004 & 3.6 \\ 
\midrule
Ours (normal GAN) & 0.0015 & 0.0060 \\
\textbf{Ours (FM GAN)} & 0.0014 & \textbf{0.0020} \\
Ours (bad-GAN) & \textbf{0.0012} & 0.0052 \\ 
\bottomrule
\end{tabular}
\label{tab2}
\end{table}

\begin{figure}[ht!]
 \centering
 \mbox{
 \subfigure[U-Net]{\includegraphics[height=0.35\linewidth]{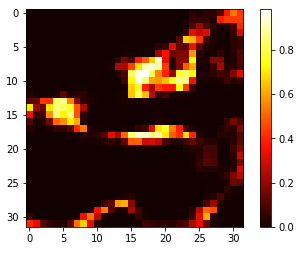}\label{fig71}}
 
 \hspace{1mm}
 
 \subfigure[Our model]
 {\includegraphics[height=0.35\linewidth]{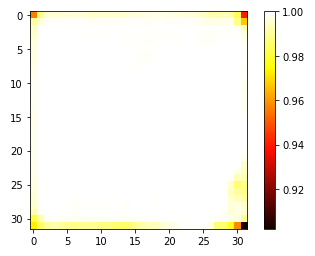}\label{fig72}}
 }
 \caption{Fake class probability predicted by U-Net and our FM GAN model for an input fake patch. Note that patches are 3D, and a single 2D slice is shown here for visualization purposes.} 
 \label{fig7}
\end{figure}

To better asses the impact on segmentation of adding unlabeled and generated images, Table \ref{tab2} gives the mean unlabeled and fake loss of the discriminator computed over test data. For this experiment, we extracted labeled patches from test images and generated an equal number of fake patches with the different GAN models. The high fake loss value of simple U-Net confirms that this model cannot discriminate between real and fake data. This limitation of U-Net can also been seen in Fig. \ref{fig7}, which gives the predicted probabilities of U-Net and our FM GAN model for a fake input patch. Unlike U-Net, the proposed model gives a fake class probability near to 1 (i.e., white color) for all voxels of the patch.   

\begin{figure*}[ht!]
 \centering
 \includegraphics[width=140mm]{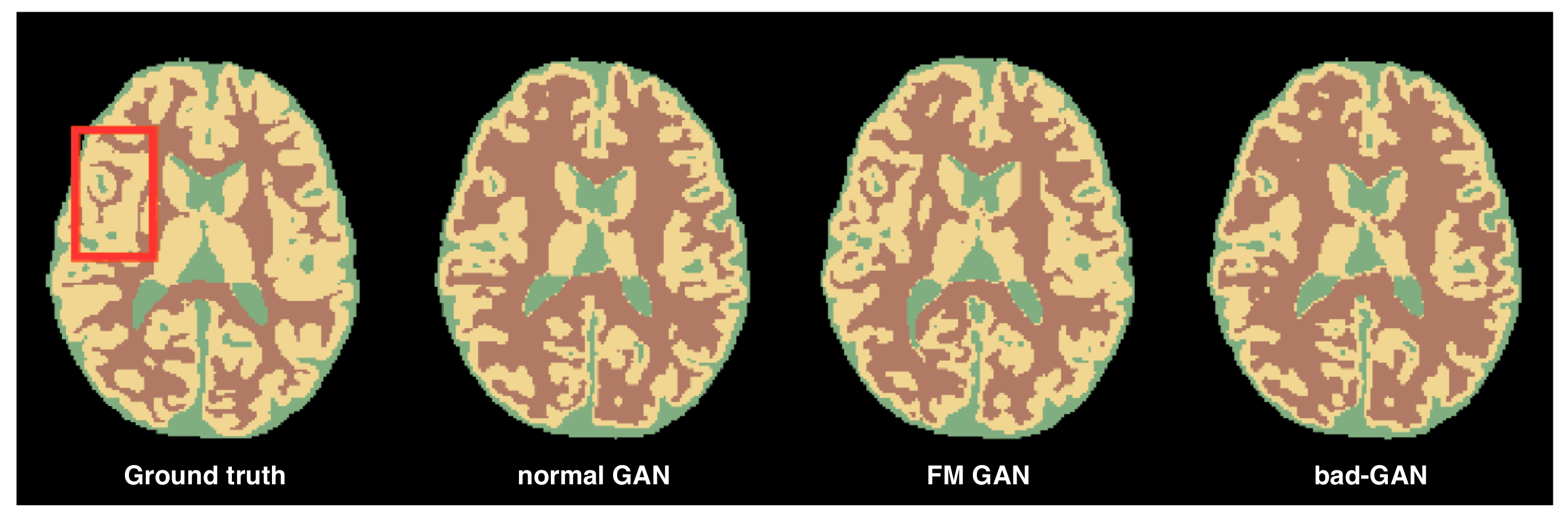}
 \caption{Segmentation of Subject 10 of the iSEG-2017 dataset predicted by different GAN-based models, when trained with 2 labeled images. The red box highlights a region in the ground truth where all these models give noticeable differences.}
 \label{fig8}
\end{figure*} 

\begin{table*}[ht!]
\centering
\caption{DSC and ASD (mm) results on 3 test images and 1 training image from the MRBrains 2013 dataset. Best results are highlighted in bold.}
\renewcommand{\arraystretch}{1.3}
\begin{tabular}{lcllllllc}
\toprule
\multicolumn{1}{l}{\multirow{2}{*}{\textbf{Method}}} & \multicolumn{2}{c}{\textbf{WM}} &  & \multicolumn{2}{c}{\textbf{GM}} &  & \multicolumn{2}{c}{\textbf{CSF}} \\ \cmidrule{2-3} \cmidrule{5-6} \cmidrule{8-9} 
\multicolumn{1}{c}{} & \textbf{DSC} & \textbf{ASD} &  & \textbf{DSC} & \textbf{ASD} &  & \textbf{DSC} & \textbf{ASD} \\ \midrule\midrule
U-Net & 0.66 & 1.78 &  & 0.67 & 1.75 &  & 0.44 & 3.30 \\
\textbf{Ours (FM GAN)} & \textbf{0.75} & \textbf{0.96} &  & \textbf{0.72} & \textbf{1.10} &  & \textbf{0.55} & \textbf{2.04} \\ 
\bottomrule
\end{tabular}
\label{tablast}
\end{table*}

\begin{figure}[h!]
 \centering
 \includegraphics[width=0.75\linewidth]{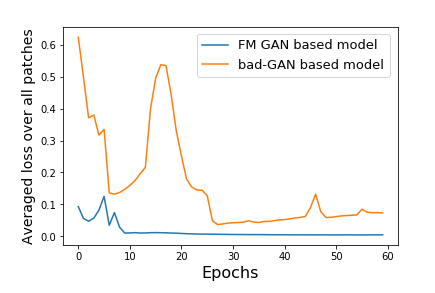}
 \caption{Feature Matching loss of bad-GAN and FM GAN models, measured at different training epochs.}
 \label{fig4}
\end{figure}

Our results indicate that FM GAN outperforms the more complex bad-GAN model (see Table \ref{tab1} and Fig. \ref{fig8}), which also adds an entropy term to have a more diverse distribution of generated examples. For  bad-GAN, we only incorporated the variational inference (VI) loss, as the low density enforcement term was not relevant in our setting given the poor sample quality. It was found that adding the VI term \cite{NIPS2017_7229} does not improve the performance of FM GAN for semi-supervised 3D image segmentation.
One possible explanation for this is the poor sample quality, which is further aggravated when increasing the entropy. 

Figure \ref{fig4} plots the feature matching loss for both the FM GAN and bad-GAN models. It can be seen that the feature matching loss of FM GAN converges quickly and remains less than that of bad-GAN, indicating a better sample generation. Patches generated by bad-GAN have a higher chance of being far from the true distribution and, hence, we may fail to learn a discriminator with a tight boundary of the true manifold. For example, there might be generated patches which are outside the true manifold but classified as true by the discriminator. This can also be seen in Table \ref{tab2}, where the average fake loss of bad-GAN is greater than that of FM GAN. Overall, the fake loss has an important contribution to performance in semi-supervised segmentation. It should produce samples that are different from true unlabeled images, while remaining close enough so that the discriminator learns useful information. 




\subsection{Validation on MR Brains dataset}

To validate our results, we also ran similar experiments on the MR Brains dataset using just 1 training example, the results of which are listed in Table \ref{tablast}. As in previous experiments, we see that the proposed technique outperforms standard U-Net, with DSC improvements of 13.6\% for WM, 7.5\% for GM, and 34\% for CSF. Likewise, our technique also yields a significant reduction in ASD: 46\% for WM, 37\% for GM, and 38\% for CSF. These results suggest the usefulness of our method for across different 3D multi-modal segmentation tasks. 

\section{Conclusion}

We presented a method for segmenting 3D multi-modal images, which can achieve performances comparable to full-supervision with only a few training samples. We showed how the method uses unlabeled data to prevent over-fitting, by learning to discriminate between true and generated fake patches. The proposed model can be employed to enhance any segmentation network in a low data setting, where the network fails to produce a good segmentation output. It also provides a new technique for few-shot learning, obviating the need for an initial pre-trained network by leveraging the semi-supervised learning ability of GANs. Moreover, results on the iSEG-2017 and MRBrains 2013 datasets showed our method's potential for reducing the burden of acquiring annotated medical data.

Our experiments explored different generator losses and their impact on segmentation performance. We showed empirically that FM GAN performs better than bad-GAN for segmenting 3D multi-modal brain MRI images. Our method can be extended to other 3D multi-modal image segmentation tasks with any state-of-the-art segmentation network as discriminator.



%





\ifCLASSOPTIONcaptionsoff
 \newpage
\fi



\bibliographystyle{IEEEtran}
\bibliography{Bibliography}





\vfill


\end{document}